# Difference Attention Based Error Correction LSTM Model for Time Series Prediction

**Yuxuan Liu[1,2*], Jiangyong Duan[1] and Juan Meng[1]**

[1] Key Laboratory of Space Utilization, Technology and Engineering Center for Space Utilization, Chinese Academy of Sciences, Beijing, 100094, China

[2] University of Chinese Academy of Sciences, Beijing, 100049, China

[*]Corresponding author's e-mail: liuyuxuan17@csu.ac.cn

**Abstract.** In this paper, we propose a novel model for time series prediction in which difference-attention LSTM model and error-correction LSTM model are respectively employed and combined in a cascade way. While difference-attention LSTM model introduces a difference feature to perform attention in traditional LSTM to focus on the obvious changes in time series. Error-correction LSTM model refines the prediction error of difference-attention LSTM model to further improve the prediction accuracy. Finally, we design a training strategy to jointly train the both models simultaneously. With additional difference features and new principle learning framework, our model can improve the prediction accuracy in time series. Experiments on various time series are conducted to demonstrate the effectiveness of our method.

## 1. Introduction

Time series prediction is an important issue in machine learning and data mining communities [1], which is critical for many applications, such as in industry and spacecraft where time series are predicted to indicate the health of the equipment. However, the complexity of time series makes the time series prediction challenging.

Traditional time series prediction methods are divided into three categories: statistical methods, machine learning methods and deep learning methods. Statistical methods, such as exponential smoothing, Auto Regressive Integrated Moving Average [2], and Kalman Filter [3], rely on the selection of parameter models. However, it is hard to know the distribution of time series in advance, which constrains the application of statistical methods. Machine learning methods, such as Support Vector Machine [4] and Gaussian Process Regression [5], manually extract features and then learn the relevant patterns, where feature extraction requires additional expert knowledge for them.

Recently, deep learning methods have been proposed to automatically extract useful features instead of hand-crafted features to predict time series [6]. Among deep learning methods, LSTM can learn the long-term dependence of time series and are widely applied in time series prediction [7], such as GRU [8], Dual-Memory LSTM [9], LSTM based on evolutionary attention mechanism [10], Phased LSTM [11] and Auto-Encoder Based LSTM [12]. These methods either simplify the LSTM model to improve the prediction speed, or improve the LSTM from the perspective of multidimensional time series prediction.

Observing that there are usually abrupt changes in time series which may lead to low prediction accuracy, we propose a new DAEC-LSTM model for time series prediction to explicitly model the changes in the time series. Specifically, an attention mechanism based on the time series difference is employed in our difference-attention LSTM model to enforce our model focusing on the data changes. Then, error correction is carried out to refine the prediction errors of difference-attention LSTM by an error-correction LSTM model. Finally, by combining difference-attention LSTM and error-correction LSTM in a cascade way, we propose a training method to jointly learn them. Our method can efficiently learn the time series tendency of time series, especially in the abrupt change positions.

## 2. Our model

Time series prediction is difficult, especially in the position where time series changes abruptly. While time series difference can characterize the regularity of time series changes, it will be quite useful to introduce time difference as a feature for prediction. To introduce the time difference in LSTM model, we propose a new DAEC-LSTM model for time series prediction.

Our model is illustrated in Fig. 1. It consists of two models, i.e., difference-attention LSTM (DA-LSTM) model and error-correction LSTM (EC-LSTM model). While difference-attention LSTM introduces difference features with an attentional mechanism to learn the time series changes, error-correction LSTM further refines the errors in difference-attention LSTM model. Finally, we employ a training strategy to jointly train the both models simultaneously. The details of our models are described below.

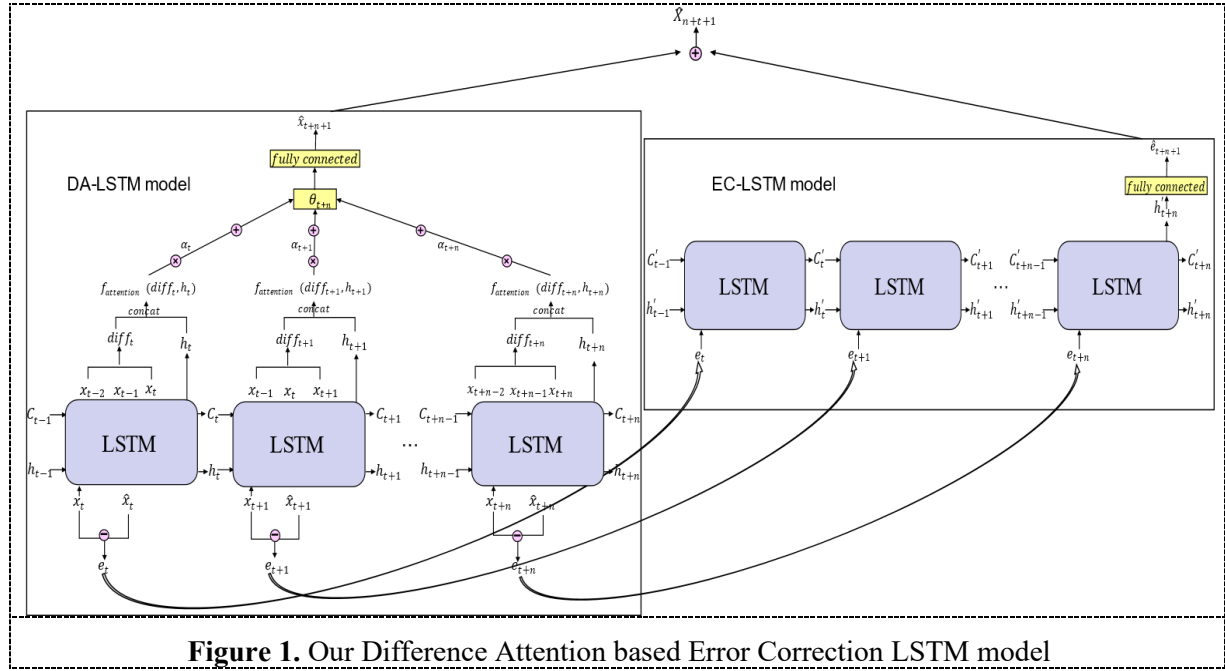

**Figure 1.** Our Difference Attention based Error Correction LSTM model

### 2.1. Difference Attention-LSTM model

In our difference attention-LSTM (DA-LSTM) model, we employ a new attentional mechanism in LSTM model to enforce our model focusing on the time series changes.

We use standard LSTM [13] as the basic model which is stacked by sequential LSTM cells as shown in Fig. 2. A typical LSTM cell is configured mainly by three gates, i.e., input gate $i_t$, forget gate $f_t$ and output gate $O_t$, to update the cell state $C_t$ and output hidden state $h_t$. Let us denote the input as $X = (x_1,...,x_n)$, hidden state of memory cells as $H = (h_1,...,h_n)$, output time series as $\hat{X} = (\hat{X}_1,...,\hat{X}_n)$. LSTM cell updates its state as follows:

$$f_t = \sigma(W_f \cdot [h_{t-1}, x_t]) + b_f \tag{1}$$

$$i_t = \sigma(W_i \cdot [h_{t-1}, x_t]) + b_i \tag{2}$$

$$\tilde{C}_t = \tanh(W_C \cdot [h_{t-1}, x_t]) + b_C \tag{3}$$

$$C_t = f_t * C_{t-1} + i_t * \tilde{C}_t \tag{4}$$

$$O_t = \sigma(W_O \cdot [h_{t-1}, x_t]) + b_O \tag{5}$$

$$h_t = O_C * \tanh(C_t) \tag{6}$$

where weight matrices are denoted as W, and bias vectors denoted as b.

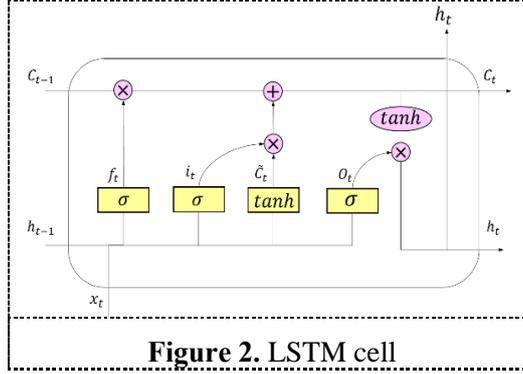

**Figure 2.** LSTM cell

Different from standard LSTM model, which only uses the last time step's hidden state to make prediction, our DA-LSTM model takes all the information at each time step in attention mechanism and employs additional magnitude variation features to focus on the time series changes. The detail of DA-LSTM is shown as Fig. 3.

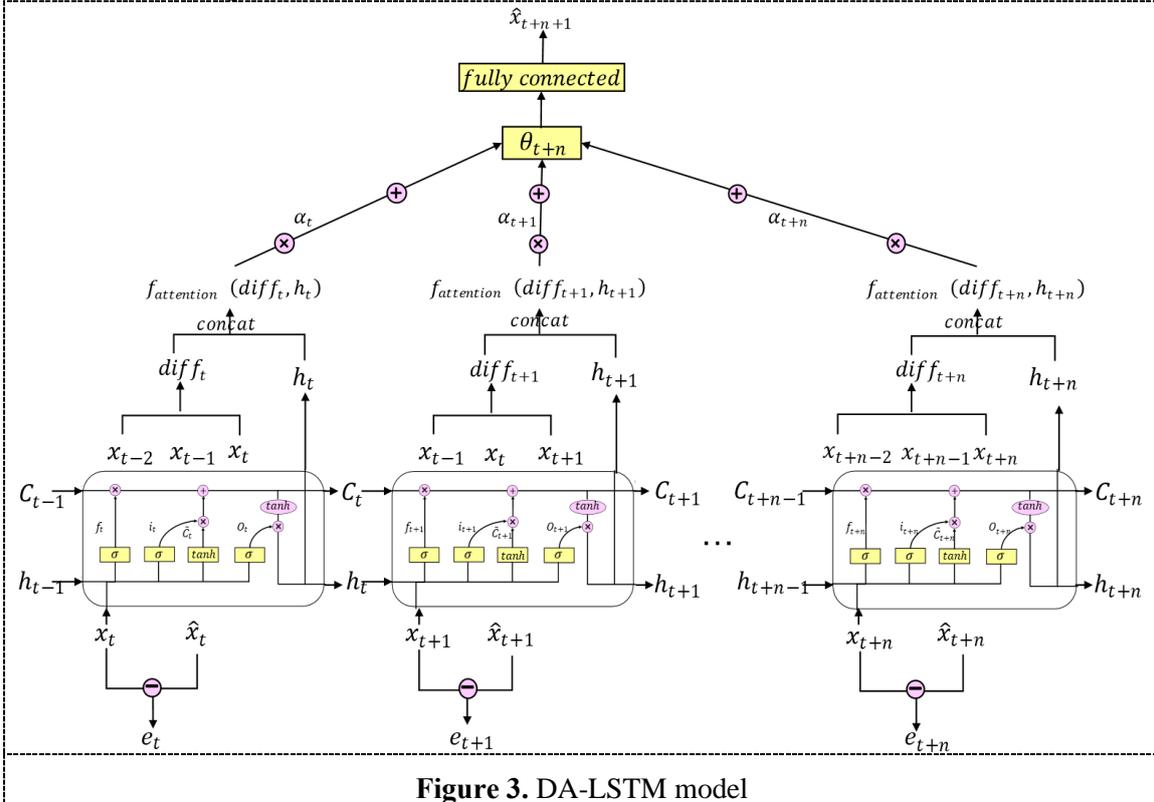

**Figure 3.** DA-LSTM model

Let $X_{t:t+n} = (x_t, ..., x_{t+n})$ be the input time series in our model, $\hat{x}_{t+n+1}$ be the output, $H_{t-1:t+n} = (h_{t-1}, ..., h_{t+n})$ be the hidden state, $C_{t-1:t+n} = (C_{t-1}, ..., C_{t+n})$ be the cell unit state. In DA-LSTM, we extract the difference feature $diff_t = [(x_{t-1} - x_{t-2})^2, (x_t - x_{t-1})^2]$ to measure the input time series variation. By concatenating the hidden state and the difference feature into a new hidden feature

$\tilde{h}_t = [diff_t, h_t]$, the attentional weight $\alpha_t$ indicating the importance of each time step is estimated by

$$\alpha_t = \frac{\exp\left(f_{attention}\left(W, [diff_t, h_t]\right)\right)}{\sum_{k=t}^{t+n} \exp\left(f_{attention}\left(W, [diff_k, h_k]\right)\right)} \quad (7)$$

Given $\alpha_t$ and the hidden state $\tilde{h}_t$, we replace the output $\hat{X}_{t+n+1} = f_{fc}(h_{t+n})$ computed by the full connection with the hidden state of the last cell in standard LSTM by the new formulation $\hat{x}_{t+n+1} = f_{fc}(\theta_{t+n})$ where $\theta_{t+n} = \sum_{i=t}^{t+n} \alpha_i \tilde{h}_i$ is the weighted sum of the concatenated hidden layer $\tilde{h}_i$.

### 2.2. Error Correction-LSTM model
In addition to DA-LSTM model, we also propose error-correction LSTM (EC-LSTM) model to predict the error between the true value and the predicted value of the DA-LSTM model, so that it can learn data rules and patterns that DA-LSTM has not learned. The detail is shown as Fig. 4.

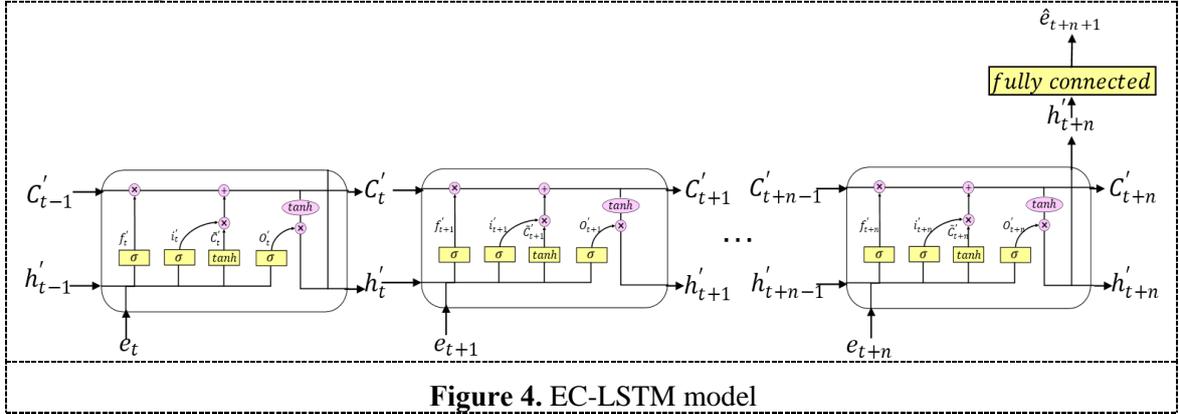

**Figure 4.** EC-LSTM model

EC-LSTM model is similar to the standard LSTM model except that the input of EC-LSTM is the error between the predicted output of DA-LSTM and the input time series. Specifically, the error time series $E_{t:t+n}$ between the predicted output value of DA-LSTM and the input time series is defined as

$$E_{t:t+n} = (e_t, \ldots e_{t+n}) = (x_t - \hat{x}_t, \ldots x_{t+n} - \hat{x}_{t+n}) \quad (8)$$

Where $\hat{x}_t = f_{DA}(x_{t-1}, x_{t-2}, x_{t-3}, \ldots x_{t-n})$ is the prediction of DA-LSTM model.

Based on the error time series $E_{t:t+n}$ in the previous time steps, EC-LSTM learns and predicts the error time series due to DA-LSTM in a traditional LSTM model in the follow equation:

$$\hat{e}_{t+n+1} = f_{EC}\left((x_{t+n} - \hat{x}_{t+n}), (x_{t+n-1} - \hat{x}_{t+n-1}), \ldots (x_t - \hat{x}_t)\right) \quad (9)$$

Where $f_{EC}$ represents basic LSTM model as illustrated in [13].

Given DA-LSTM model and EC-LSTM model, we combine them to produce our difference attention based error correction LSTM Model to improve the prediction.

### 2.3. Difference Attention based Error Correction LSTM Model
Our difference attention based error correction LSTM (DAEC-LSTM) Model combines DA-LSTM and EC-LSTM in a cascade way. While DA-LSTM extracts the information in the original data, EC-LSTM learns the data pattern that was not grasped by the DA-LSTM model to refine the prediction.

Our DAEC-LSTM model is defined as

$$\hat{X}_{t+n+1} = f_{DA}(X_{t:t+n}) + f_{EC}(E_{t:t+n}) \quad (10)$$

where $f_{DA}(X_{t:t+n})$ and $f_{EC}(E_{t:t+n})$ represent DA-LSTM and EC-LSTM model respectively.

To train DAEC-LSTM model, we first train the model parameters of $f_{DA}(X_{t:t+n})$ and $f_{EC}(E_{t:t+n})$ respectively. For $f_{DA}(X_{t:t+n})$, we use the loss $Loss_1 = (x_{t+n+1} - f_{DA}(X_{t:t+n}))^2$ and use Adam to optimize it [14]. For $f_{EC}(E_{t:t+n})$, we use the loss $Loss_2 = (e_{t+n+1} - f_{EC}(E_{t:t+n}))^2$, and employ Adam method to optimize it. Then the learned parameters of $f_{DA}(X_{t:t+n})$ and $f_{EC}(E_{t:t+n})$ are used as an initialization for our entire model in (10), and use the loss $Loss = (x_{t+n+1} - \hat{X}_{t+n+1})^2$, and optimize it also by Adam.

## 3. Experiments

We demonstrate our method on three temperature telemetry datasets for the space payload. Our DAEC-LSTM model is compared with standard LSTM [13], Bi-directional LSTM (Bi-LSTM) [15], stacked LSTM (S-LSTM) [16], and also compared with DA-LSTM and EC-LSTM model. For all these LSTM methods, the input sequence length and hidden size are set 40 and 100 respectively in the experiments. For stacked LSTM, we set 3 for the number of stacked layers. In comparison, mean square error (MSE), mean absolute percentage error (MAPE) are used as evaluation criteria, which are defined as equation (11-12). The smaller MSE and MAPE, the higher prediction accuracy is obtained.

$$MSE = \frac{1}{N} \sum_{i=1}^{N} (x_i - \hat{x}_i) \qquad (11)$$

$$RMSE = \sqrt{\frac{1}{N} \sum_{i=1}^{N} (x_i - \hat{x}_i)^2} \qquad (12)$$

Where $\hat{x}_i$ and $x_i$ represent the predicted and ground truth value.

Table 1 shows the performance comparison results of the construction of the prediction model. It can be seen that our DAEC-LSTM model obtains the highest prediction accuracy. For LSTM to model complex time series, it cannot fully grasp the patterns and variations of data. If more time steps are used in the experiment, the ability of the model to pay attention to the characteristics of sub-windows is far from enough. In addition, in the position existing large fluctuations and variations, the standard LSTM prediction results do not match the original data well, so the accurate prediction cannot be achieved. For Bi-directional LSTM makes full use of the context relationship between the forward and backward time directions of the time series. The output of the current time is not only related to the previous state, but also may be related to the future state, so as to learn the information with long-term dependence on time. But compared with DAEC-LSTM, Bi-LSTM only makes the information before and after time steps more balanced, giving each time step the same degree of importance, without highlighting the importance of a certain time step information. The stacked LSTM improves the efficiency of training and achieves higher accuracy only by increasing the depth of LSTM network, which obtain less information than DAEC-LSTM. Therefore, by introducing the attention mechanism, DA-LSTM introduces magnitude variation and focuses on the part of fluctuation jump in time series.

Although the prediction effect of DA-LSTM is better than that of the prediction model using only the basic LSTM, it was found that there was still a lag in prediction. In some time steps where the value changes greatly, the predicted value cannot change immediately with the changing trend of the data. In DA-LSTM, though the magnitude variation is constructed for auxiliary prediction, it is still challenging to fully grasp the data rules and patterns due to the complexity of the data itself. It may be that a large part of the existing patterns in the data are smooth and uniform, and the model has learned more information of uniform changes, while the model has not learned enough about other forms of data change patterns. Therefore, an additional error correction module is added. Based on the standard LSTM, EC-LSTM mechanism is added for prediction and verify the effectiveness of error correction. Thus our DAEC-LSTM model produces satisfying results.

**Table 1.** Prediction accuracy comparison

| Dataset | Method | MSE | MAPE |
|---|---|---|---|
| Dataset 1 | LSTM | 0.0034 | 0.22% |
| | Bi-LSTM | 0.0025 | 0.20% |
| | S-LSTM | 0.0023 | 0.19% |
| | DA-LSTM | 0.0019 | 0.18% |
| | EC-LSTM | 0.0018 | 0.16% |
| | DAEC-LSTM | **0.0007** | **0.10%** |
| Dataset 2 | LSTM | 0.0151 | 2.68% |
| | Bi-LSTM | 0.0146 | 2.60% |
| | S-LSTM | 0.0145 | 2.58% |
| | DA-LSTM | 0.0139 | 2.47% |
| | EC-LSTM | 0.0143 | 2.53% |
| | DAEC-LSTM | **0.0126** | **2.31%** |
| Dataset 3 | LSTM | 0.0061 | 0.30% |
| | Bi-LSTM | 0.0059 | 0.27% |
| | S-LSTM | 0.0057 | 0.26% |
| | DA-LSTM | 0.0051 | 0.23% |
| | EC-LSTM | 0.0053 | 0.25% |
| | DAEC-LSTM | **0.0046** | **0.19%** |

The comparison between the ground truth and predicted time series of our method is illustrated in Fig. 5. Our model produces the prediction result which is quite similar to the ground truth. Even in the positions of abrupt changes, we can still produce satisfying results.

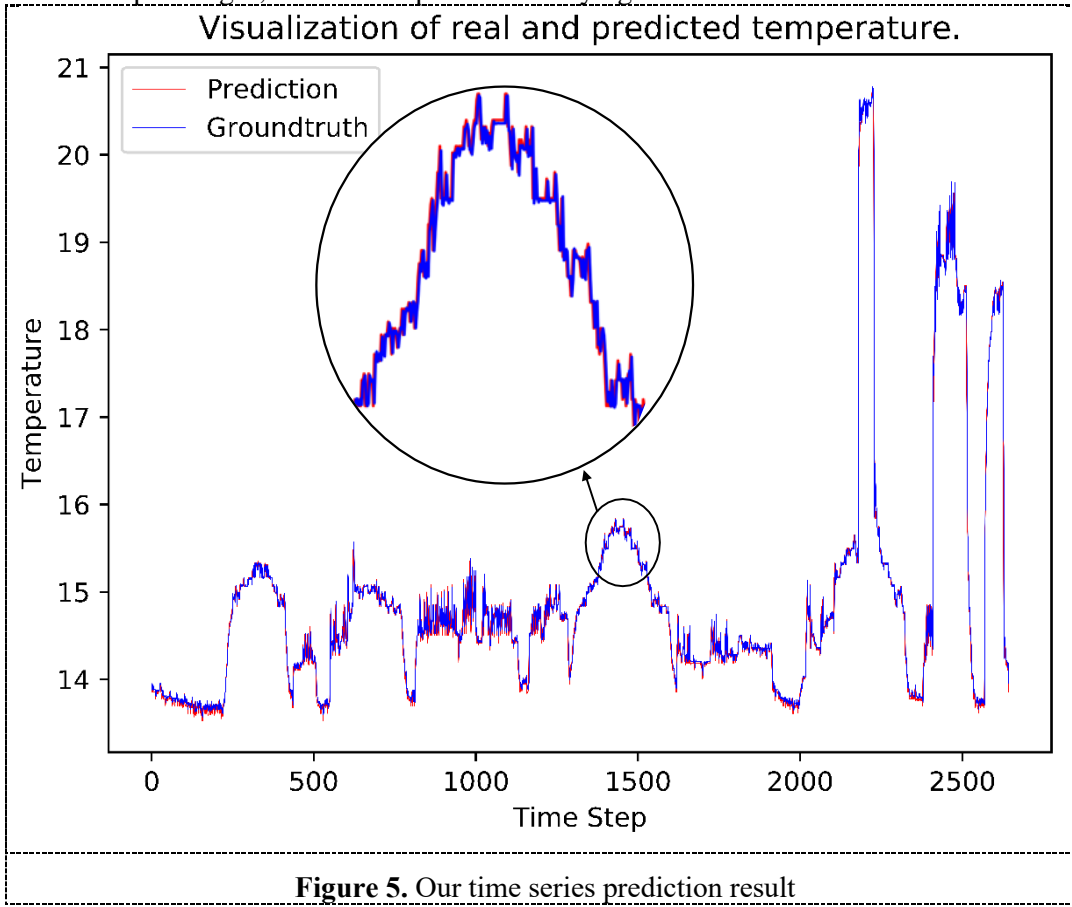

**Figure 5.** Our time series prediction result

## 4. Conclusion
This paper proposed a novel DAEC-LSTM model for time series prediction. By combing the difference attention mechanism and error correction mechanism in a principled framework, our model produced satisfying results for complex time series with abrupt changes. The experiments demonstrated the effectiveness of our method.

**Acknowledgments.** This work is supported by the National Natural Science Foundation of China (No. 61901454, No. 61971404 and No. 61501434)